\def\year{2022}\relax
\documentclass[letterpaper]{article} 
\usepackage{amsmath}
\usepackage{multirow}
\usepackage{booktabs}
\usepackage{xcolor}
\usepackage{makecell} 
\usepackage{amsfonts}
\usepackage{nameref}
\usepackage{aaai22}  
\usepackage{times}  
\usepackage{helvet}  
\usepackage{courier}  
\usepackage[hyphens]{url}  
\usepackage{graphicx} 
\usepackage{natbib}  
\usepackage{caption} 
\DeclareCaptionStyle{ruled}{labelfont=normalfont,labelsep=colon,strut=off} 
\usepackage{algorithm}
\usepackage{algorithmic}
\usepackage{hyperref}
\usepackage[normalem]{ulem}
\usepackage{xcolor}

\usepackage{newfloat}
\usepackage{listings}
\floatstyle{ruled}
\newfloat{listing}{tb}{lst}{}
\floatname{listing}{Listing}
\title{MoGaFace: Momentum-Guided and Texture-Aware Gaussian Avatars for Consistent Facial Geometry}
\author{
    Yujian Liu\textsuperscript{\rm 1, \rm 2},
    Linlang Cao\textsuperscript{\rm 1},
    Chuang Chen\textsuperscript{\rm 1},
    Fanyu Geng \textsuperscript{\rm 1},\\
    Dongxu Shen\textsuperscript{\rm 3}, 
    Peng Cao\textsuperscript{\rm 4}, 
    Shidang Xu\textsuperscript{\rm 2}\footnotemark[1],
    Xiaoli Liu\textsuperscript{\rm 1}\thanks{Corresponding author.}
}
\affiliations{
    \textsuperscript{\rm 1}AiShiWeiLai AI Research, Beijing, China\\
    \textsuperscript{\rm 2}South China University of Technology, Guangzhou, China\\
    \textsuperscript{\rm 3}Hong Kong University of Science and Technology (Guangzhou), Guangzhou, China\\
    \textsuperscript{\rm 4}Northeastern University, Shenyang, China\\

}

\usepackage{bibentry}

\begin{document}

\maketitle

\begin{abstract}
Existing 3D head avatar reconstruction methods adopt a two-stage process, relying on tracked FLAME meshes derived from facial landmarks, followed by Gaussian-based rendering. However, misalignment between the estimated mesh and target images often leads to suboptimal rendering quality and loss of fine visual details. In this paper, we present MoGaFace, a novel 3D head avatar modeling framework that continuously refines facial geometry and texture attributes throughout the Gaussian rendering process. To address the misalignment between estimated FLAME meshes and target images, we introduce the Momentum-Guided Consistent Geometry module, which incorporates a momentum-updated expression bank and an expression-aware correction mechanism to ensure temporal and multi-view consistency. Additionally, we propose Latent Texture Attention, which encodes compact multi-view features into head-aware representations, enabling geometry-aware texture refinement via integration into Gaussians. Extensive experiments show that MoGaFace achieves high-fidelity head avatar reconstruction and significantly improves novel-view synthesis quality, even under inaccurate mesh initialization and unconstrained real-world scenarios. 

\end{abstract}


\section{Introduction}
Creating animatable head avatars has long been a challenge in computer vision \cite{beeler2010high,thies2016face2face,xu2023latentavatar}. Photorealistic and dynamic rendering from arbitrary viewpoints is essential for applications such as gaming, film production, immersive telepresence, and AR/VR. Equally important is the ability to control these avatars and ensure they generalize well to novel poses and expressions.

Neural Radiance Fields (NeRF) \cite{mildenhall2021nerf} and its variants \cite{barron2021mip, chen2022tensorf,muller2022instant} have achieved remarkable results in reconstructing static scenes. Subsequent works have extended NeRF to dynamic scenarios, such as modeling arbitrary motions and generating human-specific head or body avatars \cite{peng2024synctalk, liu2025syncanimation}. Despite synthesizing high-quality novel views \cite{lombardi2021mixture, gao2022reconstructing}, NeRF-based methods suffer from limited controllability and generalization due to implicit representations \cite{yu2024gaussiantalker,zhang2023explicifying}, and their costly volumetric rendering prevents real-time applications \cite{li2024talkinggaussian}.
\begin{figure}[t]
\centering
\includegraphics[width=1\columnwidth]{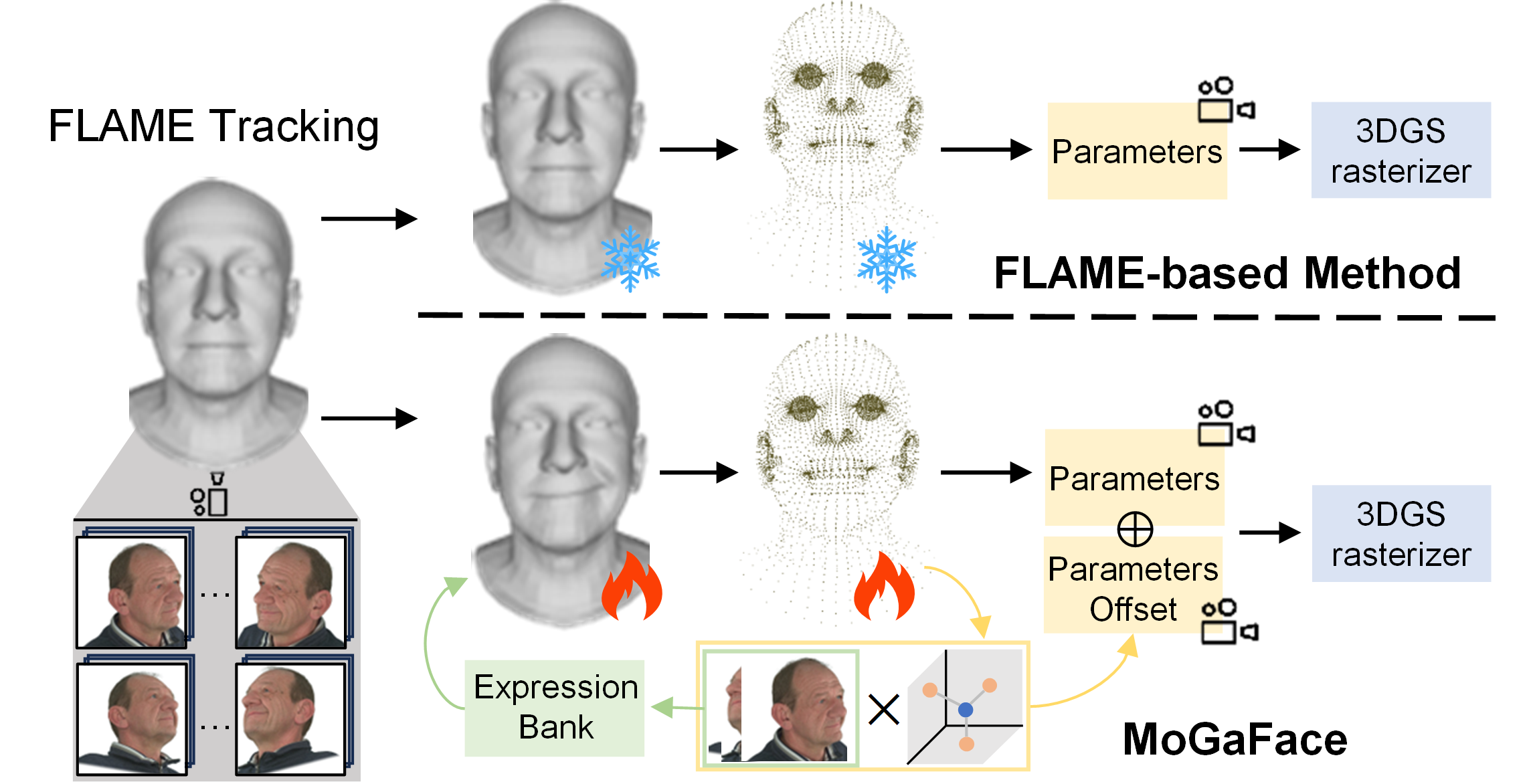} %
\caption{MoGaFace introduces a novel approach that jointly optimizes facial expression coefficients and texture attributes, enabling high-fidelity facial texture rendering based on precise alignment with the FLAME geometry.}
    \label{fig:figure1}
\end{figure}

Recently, 3D Gaussian Splatting (3DGS) \cite{kerbl20233d} has gained widespread attention for achieving high-quality rendering and real-time performance in novel view synthesis. Unlike NeRF, 3DGS models surfaces with independent 3D Gaussians defined by opacity, anisotropic covariance, and spherical harmonics \cite{tang20243igs, kwon2024generalizable, xie2025envgs}. By optimizing these discrete geometric primitives throughout the scene, 3DGS not only surpasses NeRF in rendering fidelity but also supports real-time applications \cite{liu2024citygaussian}. Building on this, recent studies \cite{qian2024gaussianavatars, wang20253d, aneja2024gaussianspeech,wang20253d} have explored the integration of 3DGS with FLAME \cite{li2017learning}, aiming to leverage the parametric nature of FLAME to enable explicit and temporally consistent control over facial expressions and head poses. However, current FLAME-based 3DGS approaches commonly follow a two-stage pipeline, where unsupervised 2D-to-3D mesh estimation is first performed, followed by mesh-to-image gaussian splatting, as shown in Fig.\ref{fig:figure1}(top). The second stage generally relies on the quality of mesh estimation through multi-view tracking, which limits scalability in lightweight scenarios where camera viewpoints are unavailable \cite{zheng2024gps,wei2025pcr}. Moreover, mesh-based texture representations lack image-space constraints, making photorealistic view synthesis and controllable expression modeling challenging to unify.

To address these limitations, we propose MoGaFace, a 3D head avatar framework that jointly  refines geometry and texture for realistic and expressive avatar synthesis (Fig.\ref{fig:figure1}(bottom). We first propose a Momentum-Guided Consistent Geometry module to align 2D images with 3D facial geometry across temporal and multi-view settings. For temporal consistency, an Expression-Aware Dynamic Correction mechanism is introduced, which leverages multi-view images to refine FLAME expressions, aiming to better align facial geometry with image-space details at each time step. However, performing splatting independently for each view can lead to inconsistent FLAME across views at the same time.
To ensure multi-view consistency, we introduce a Momentum-updated Expression Bank that averages expression corrections across views at the same timestamp  using a momentum strategy  \cite{he2020momentum}, enforcing consistent FLAME geometry during training and inference. Finally, to establish a stronger constraint from image to mesh, we propose Latent Texture Attention, which encodes compact multi-view features into a head-aware representation. This allows for enhanced texture information to be integrated into the 3D Gaussians by supplementing geometry with texture-related cues. Experimental results demonstrate that MoGaFace not only improves the clarity and texture quality of novel-view synthesis but also enables high-fidelity 3D head avatar reconstruction without relying on known camera parameters, significantly broadening its applicability in real-world scenarios. In summary, the main contributions of our work are as follows:
\begin{itemize}
    \item We propose MoGaFace, a novel 3D head avatar modeling framework that jointly refines expression coefficients and texture attributes during the rendering process, enabling realistic and expressive avatar synthesis.
    
    \item We introduce the Momentum-Guided Consistent Geometry module, a key component that significantly enhances the consistency and accuracy of FLAME expression modeling. By refining expressions from multi-view images and propagating corrections via momentum-based updates, it ensures coherent geometry across both time and viewpoints.
    
    \item We propose Latent Texture Attention, which encodes compact multi-view features into head-aware representations, allowing texture-related cues to be effectively integrated into 3D Gaussians for enhanced texture quality.

    \item Extensive experiments demonstrate that MoGaFace achieves high-fidelity 3D head avatar reconstruction and significantly improves the quality of novel-view synthesis, outperforming existing state-of-the-art methods on multiple benchmarks.
\end{itemize}

\section{Related Work}

\paragraph{Monocular 3D Gaussian Splatting.}
Early 3D head–avatar reconstruction methods employed implicit neural radiance fields (NeRF) \cite{mildenhall2021nerf}, fitting dense MLPs in 4D space to achieve view-consistent volume rendering. Recent studies turn to monocular 3D Gaussian Splatting \cite{fei20243d, qiu2025anigs, ma20243d}, which discretises the radiance field into anisotropic 3D Gaussians and rasterises them for real-time synthesis.  
MonoGaussianAvatar \cite{chen2024monogaussianavatar} couples learnable Gaussians with a deformation field, directly optimising Gaussian parameters from a single-view video.  
Gaussian Blendshapes \cite{ma20243d} transfers the classic mesh-blendshape paradigm to Gaussian space, linearly mixing Gaussian blendshapes for real-time driving. 
GaussianHead \cite{wang2025gaussianhead} introduces a learnable Gaussian derivation mechanism to alleviate texture sparsity. Monocular reconstruction is fundamentally flawed \cite{arampatzakis2023monocular}, as a single 2D view cannot fully recover 3D structure, leading existing monocular Gaussian splatting methods to lack both depth information and multi-view constraints \cite{liu2025review}.

\paragraph{Multi-view 3D Gaussian Splatting.}
In multi-camera or turn-table settings, multi-view 3D Gaussian Splatting fully exploits stereo cues, delivering faithful appearance and accurate geometry \cite{chen2024mvsplat, liu2024mvsgaussian}. Gaussian Avatars \cite{qian2024gaussianavatars} attaches Gaussians to FLAME triangles and proposes a binding-inheritance strategy. Gaussian Head Avatar \cite{xu2024gaussian} extends this framework by introducing a learnable deformation field to capture complex expressions, designed as the first 3D Gaussian-based avatar model for high-resolution reconstruction. TensorialGaussianAvatar  \cite{wang20253d} improves memory and runtime efficiency by encoding static appearance into tri-planes and representing dynamic opacity with 1D feature lines.
All the above methods adopt a two-stage rendering pipeline, rely heavily on precisely estimated meshes, and lack sufficient texture detail.

\section{Method}

\begin{figure*}[t]
\centering
\includegraphics[width=1 \textwidth]{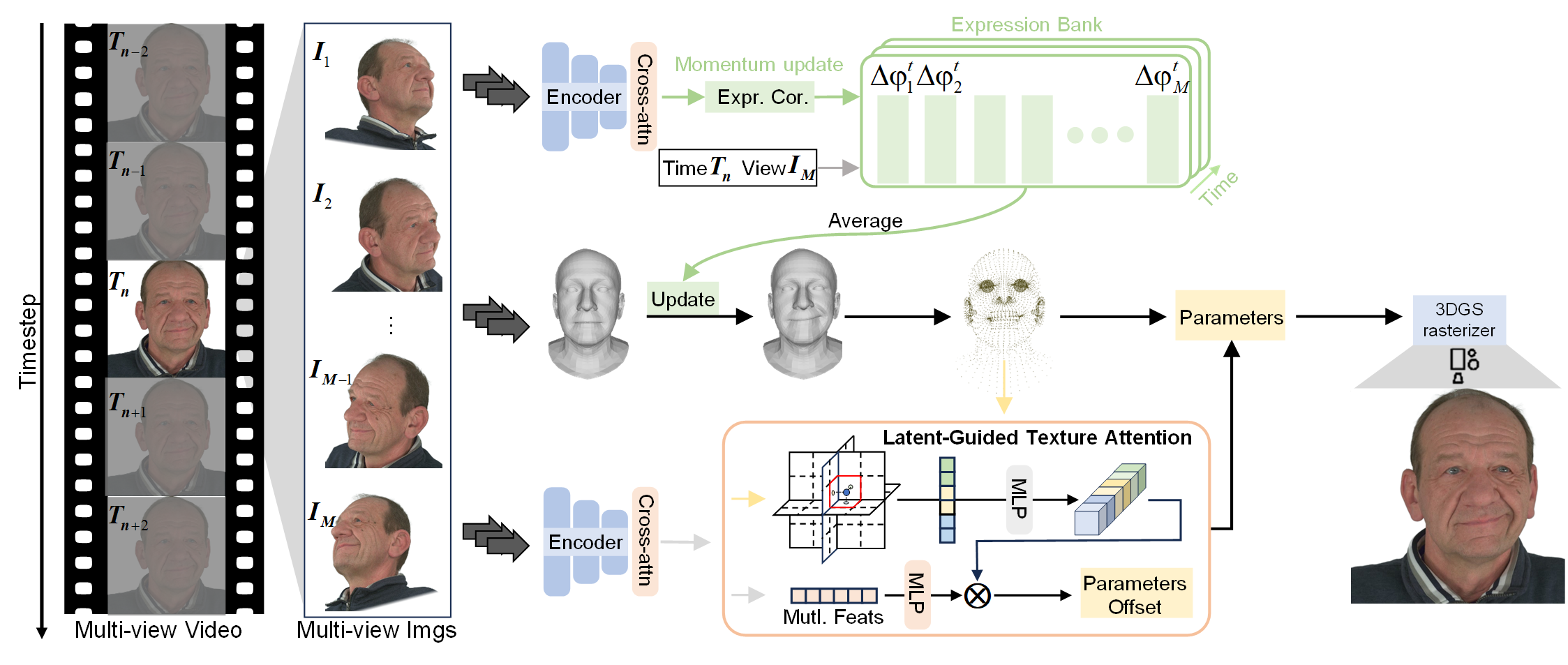} %
\caption{Framework: Given multi-view images, MoGaFace initializes FLAME via tracking, refines expressions through a Momentum-Guided Geometry module for consistent and accurate fitting, and embeds 3D Gaussian textures using a Latent Texture Attention module that exploits multi-view texture cues.}
    \label{fig:figure2}
\end{figure*}

This section introduces the proposed MoGaFace (Fig.~\ref{fig:figure2}), which consists of two key modules. The Momentum-Guided Consistent Geometry module integrates dynamic expression correction with a momentum-updated expression bank to enforce consistent, corrected FLAME meshes across views at the same timestamp during rendering. The Latent-Guided Texture Attention module adaptively encodes mesh textures by leveraging multi-view latent features, enhancing fine-grained visual details. The final part of the section outlines the loss functions and training strategy used to enable high-fidelity dynamic head reconstruction.
\subsection{Preliminaries}
\label{sec:preliminaries}

\textbf{3D Gaussian Splatting.} 3D Gaussian Splatting (3DGS)\cite{kerbl20233d} is a recent explicit rendering technique that enables real-time novel view synthesis from multi-view images and known camera parameters. A scene is represented as a set of 3D Gaussian splats, each parameterized by position $\mu \in \mathbb{R}^3$, scale $s \in \mathbb{R}^3$, rotation quaternion $q \in \mathbb{R}^4$, opacity $\alpha \in \mathbb{R}$, and view-dependent color encoded via spherical harmonics coefficients $\text{SH} \in \mathbb{R}^{(k+1)^2 \times 3}$. These Gaussians attributes are projected into the image plane using camera intrinsics and extrinsics, and rendered via a differentiable alpha-blending rasterizer. The final pixel colors are computed by depth-aware sorting and blending of the projected Gaussians.

\noindent\textbf{FLAME-based Avatar Modeling.} To extend 3D Gaussian Splatting to human head modeling with controllable expressions and poses, representative methods such as Gaussian Avatars\cite{qian2024gaussianavatars} introduce a binding mechanism between each Gaussian primitive and the FLAME mesh, a parametric 3D facial model. Given multi-view facial videos and known camera parameters, a common approach to obtaining per-frame FLAME parameters in head avatar modeling is to employ landmark-based multi-view tracking algorithms, such as VHAP. These methods typically rely on detected 2D facial landmarks across views, combined with camera intrinsics and extrinsics, to optimize the FLAME model through multi-view landmark reprojection loss. In this setting, the FLAME model is parameterized by four components: global rigid motion $\in \mathbb{R}^6$, joint-based articulation $\in \mathbb{R}^{15}$, shape $\in \mathbb{R}^{300}$, and expression $\in \mathbb{R}^{100}$. Non-rigid facial deformations are captured by the expression parameters, which control a linear combination of blendshapes derived from 4D facial scan data. These parameters allow the model to generate a wide range of identity-independent expressions, such as smiling, blinking, and mouth movements. While these parameters provide a strong initialization, they often fail to capture fine-grained expression dynamics when projected into image space—particularly in the absence of accurate camera parameters or under challenging conditions such as complex lighting, partial occlusion, or identity-expression entanglement. These limitations significantly hinder the applicability of such methods in lightweight or mobile deployment scenarios, where precise calibration is often unavailable.

\subsection{Momentum-Guided Consistent Geometry}
\label{sec:Momentum-Guided Consistent Geometry}
By integrating Expression-Aware Dynamic Correction and Momentum-Guided Expression Representation, our MoGaFace effectively enforces facial geometry consistency by dynamically refining per-view expression parameters and maintaining a shared, coherent FLAME mesh across views at the same timestamp.

\noindent\textbf{Expression-Aware Dynamic Correction}
To address the misalignment between the predicted mesh and the input images, we introduce a Dynamic Facial Expression-Aware Correction method, which refines the expression parameters $ \boldsymbol{\psi}^{(t)} $ during the rendering process by leveraging multi-view image supervision at each time step $t$. As illustrated in Fig. \ref{fig:figure2}(top), MoGaFace extracts multi-view geometry-constrained features from multiple facial images at the same time during the rendering process, enabling end-to-end regression of expression correction terms. Specifically, for each frame $t$, the $i$-th view image $I_i^{(t)}$ is passed through a trainable convolutional encoder to extract geometry-constrained features. The resulting encoded feature$ F_i^{(t)}$ is denoted as:
\begin{equation}
F_i^{(t)} = \text{Encoder}(I_i^{(t)}),
\end{equation}
Considering that FLAME models the entire head, a single view is often insufficient to provide complete geometric constraints for accurate expression estimation. Therefore, for each frame $t$, all views in the training set $(I_0^{(t)}, I_1^{(t)}, \ldots, I_{N}^{(t)})$ are independently passed through a feature encoder to extract geometry-constrained features. The resulting per-view features are then fused using a cross-attention module to aggregate multi-view information and capture comprehensive geometry-aware cues. Finally, the fused feature is subsequently used to regress a correction term that refines the initial FLAME expression parameters, enabling dynamic and view-consistent expression refinement throughout the rendering process.
\begin{equation}
\Delta \boldsymbol{\psi}^{(t)} = \text{MLP}((\text{CrossAttn}(F_0^{(t)}, F_1^{(t)}, \ldots, F_{N}^{(t)})))
\end{equation}
\noindent\textbf{Momentum-Guided Expression Representation.}
In multi-view 3DGS, the training process typically treats each view image as an independent supervision unit for rendering. However, the FLAME model represents the full head geometry and assumes that all views at the same timestamp share a set of pose and expression parameters. This view-consistency constraint is often ignored when expressions are changed independently from each image, leading to inconsistencies across views of the same frame.

Inspired by self-supervised learning memory bank \cite{wu2018unsupervised} and momentum update mechanisms\cite{he2020momentum}, MoGaFace introduces a Momentum-updated Expression Bank to enforce expression consistency across different views at the same time step. The bank serves as a global memory that maintains a set of shared expression correction terms for each timestamp, ensuring that the expression refinement is view-consistent. Specifically, for each frame $t$, we maintain a memory entry $\Delta \boldsymbol{\psi}^{(t)}_{\text{bank}} \in \mathbb{R}^{N \times 100}$, which is progressively updated with the correction terms $\Delta \boldsymbol{\psi}^{(t)}_i$ regressed from individual views $i \in \{0, \ldots, N-1\}$. The update is performed using exponential moving average as follows:
\begin{equation}
\Delta \boldsymbol{\psi}^{(t)}_{\text{bank}} \leftarrow m \cdot \Delta \boldsymbol{\psi}^{(t)}_{\text{bank}} + (1 - m) \cdot \Delta \boldsymbol{\psi}^{(t)}_i
\end{equation}
where $m \in [0, 1)$ is the momentum coefficient that controls the smoothness of the update and is linearly decayed over training iterations. During training, the banked correction term $\Delta \boldsymbol{\psi}^{(t)}_{\text{bank}}$ is used to supervise and regularize the per-view regressed expressions. At each timestamp $t$, the mean of the $N$ is obtained by averaging the correction terms across the $N$ per-view corrections is adopted as the FLAME expression adjustment, thereby enforcing inter-view consistency without requiring an explicit cross-view loss.

\subsection{Latent-Guided Texture Attention}

In this framework, each triangle in the FLAME mesh serves as an attachment base for a 3D Gaussian primitive. During training, both geometric attributes (position, scale, rotation) and appearance attributes (opacity) of each Gaussian are optimized. To ensure geometric consistency, Gaussians are densely anchored to the FLAME mesh via barycentric coordinates within its associated triangles, analogous to UV-based surface parameterization. Unlike traditional UV mapping that relies on high-quality mesh scans, this approach uses only real RGB images as weak supervision, which poses challenges for recovering fine-grained texture details. 

To address this, recent advances in single-view UV estimation \cite{zielonka2022towards,zielonka2022towards,feng2021learning}  are incorporated, along with the insight that texture attributes can be represented through implicit feature encoding to preserve local texture coherence. A triplane-based scheme is applied to the corrected FLAME mesh, where each vertex position is projected onto three orthogonal 2D feature planes to obtain localized texture descriptors. The triplane representation $\mathcal{H}$ consists of three feature planes:
\begin{equation}
\mathcal{H} = \{\mathcal{H}^{xy}, \mathcal{H}^{xz}, \mathcal{H}^{yz}\} \in \mathbb{R}^{n_f \times n_f \times n_{d_1}}
\end{equation}
with spatial resolution $n_f \times n_f$ and feature dimension $n_{d_1}$. For any point $p$ in canonical space, its feature is computed by projecting it onto the axis-aligned planes:
\begin{equation}
h(p) = \mathcal{H}^{xy}(p_{xy}) \oplus \mathcal{H}^{xz}(p_{xz}) \oplus \mathcal{H}^{yz}(p_{yz})
\end{equation}
where  $\oplus$ indicates feature concatenation, and $p_{xy}, p_{xz}, p_{yz}$ are the 2D projections of point $p$ onto each respective plane. 

Inspired by single-view UV estimation, where image features are shown to encode global pose, lighting, and facial details, we introduce an attention mechanism to modulate Gaussian attributes. After computing $h(p)$, expression-aware features $\mathbf{F} \in \mathbb{R}^d$, extracted from the Dynamic Facial Expression-Aware Correction module, are fused to enhance the texture encoding with global and appearance-aware cues. For each Gaussian $i$, the fusion of the texture descriptor $h(p_i)$ and image feature $\mathbf{F}$ is formulated as:
\begin{equation}
\mathbf{v}_i = \text{Attn}(\mathbf{F}, h(p_i)),
\end{equation}
where $\text{Attn}(\cdot)$ is a lightweight convolutional attention module. The output $\mathbf{v}_i$ adaptively regulates the expression-aware attributes of the Gaussian, enabling context-dependent modulation over geometric and appearance spaces.

Following the attention modulation, the resulting attention $\mathbf{v}_i$ is further decoded through a multi-head MLP to produce residual adjustments over key Gaussian attribute subspaces. These include position offset $\Delta\boldsymbol{\mu}_i \in \mathbb{R}^3$, anisotropic scale adjustment $\Delta\mathbf{s}_i \in \mathbb{R}^3$, rotation quaternion offset $\Delta\mathbf{r}_i \in \mathbb{R}^4$, and opacity shift $\Delta\alpha_i \in \mathbb{R}$. The prediction process is formulated as:
\begin{equation}
[\Delta\boldsymbol{\mu}_i, \Delta\mathbf{s}_i, \Delta\mathbf{r}_i, \Delta\alpha_i] = \text{MLP}_{\text{head}}(\mathbf{v}_i)
\end{equation}
Each of these components is then used to modulate the corresponding Gaussian attributes from their canonical values

\subsection{Training}
\label{sec:training}
\textbf{Loss Function.}The training pipeline supervises the rendered images using a combination of an $L_1$ loss and a D-SSIM loss. The overall image loss is defined as
\begin{equation}
\mathcal{L}_{\text{rgb}} = (1 - \lambda)\mathcal{L}_1 + \lambda \mathcal{L}_{\text{D-SSIM}},
\end{equation}
with the weighting factor $\lambda = 0.2$. To improve geometric consistency during facial animation, particularly under dynamic facial expressions, two regularization terms are introduced to guide the behavior of the 3D Gaussian primitives. The position loss encourages each Gaussian to remain close the centroid of its associated triangle on the FLAME mesh, reducing misalignment and deformation. It is defined as
\begin{equation}
\mathcal{L}_{\text{position}} = \left\| \max(\mu, \epsilon_{\text{position}}) \right\|_2,
\end{equation}
where $\mu$ denotes the positional offset relative to the triangle center, and $\epsilon_{\text{position}}=1$ defines a tolerance margin. In parallel, a scaling loss is applied to regularize Gaussian size. Excessively small Gaussians may lead to unstable rendering due to high-frequency ray-splat intersections, while large ones may blur fine details. To prevent such artifacts, the scaling loss penalizes values below a specified threshold:
\begin{equation}
\mathcal{L}_{\text{scaling}} = \left\| \max(s, \epsilon_{\text{scaling}}) \right\|_2,
\end{equation}
where $s$ is the scale, and $\epsilon_{\text{scaling}} = 0.6$ sets the minimum allowable size.
The final training objective combines the image loss with both geometric regularization terms as follows:
\begin{equation}
\mathcal{L} = \mathcal{L}_{\text{rgb}} + \lambda_{\text{position}} \mathcal{L}_{\text{position}} + \lambda_{\text{scaling}} \mathcal{L}_{\text{scaling}},
\end{equation}
where $\lambda_{\text{position}} = 0.01$ and $\lambda_{\text{scaling}} = 1$. The position and scaling losses are applied only to visible Gaussians, focusing supervision on perceptually relevant regions while preserving occluded areas

\noindent\textbf{Implementation Details.} For parameter optimization, the Adam optimizer is used, and the same hyperparameters are applied across all subjects. The learning rate is set to 5e-3 for Gaussian positions and 1.7e-2 for their scales. For the FLAME parameters, including translation, joint rotation, and facial expressions, learning rates of 1e-6, 1e-5, and 1e-3 are employed for different stages of training. The model is trained for a total of 1000,000 iterations. During training, the learning rate for Gaussian positions decays exponentially and reaches 1$\%$ of its original value by the final iteration. Additional mechanisms are employed to improve convergence and visual quality. Starting from the 6,000th iteration, a binding interaction loss is activated every 2,000 steps to enforce consistency between splats and mesh. Furthermore, every 100,000 iterations, the opacities of the Gaussians are reset to avoid degenerate transparency. A photo-metric head tracker is also utilized to supervise the global FLAME parameters, including shape coefficients ($\beta$), global translation ($t$), pose ($\theta$), expression ($\psi$), and vertex offset ($\Delta v$) in the canonical FLAME space. These details collectively ensure stable, accurate, and high-fidelity dynamic avatar reconstruction during both training and animation.
\section{Experimental Results}
\subsection{Implementations}

\noindent\textbf{Dataset.} 
In our experiments, we utilize 12 subject sequences, including 10 from the NeRSemble dataset\cite{kirschstein2023nersemble} and 2 recorded by ourselves using five unsynchronized cameras. The NeRSemble recordings consist of 16 camera views covering both frontal and lateral angles. For each subject, we select 21 video sequences and downsample the frames to a resolution of 802 × 550. Participants were instructed to either perform  10 predefined facial expressions or recite 10 segments of neutral speech. The final "FREE" sequence, containing spontaneous expressions, is used for self-reenactment task. For our own recordings (approximately 5 min 10 s each), each frame is cropped and resized to 512 × 512 with a green screen background. The data for each subject is split into training and self-reenactment subsets at a 4:1 ratio.
\noindent\textbf{Evaluation Metrics.} We demonstrate the superiority of our method using several commonly adopted metrics.To evaluate image quality, we employ full-reference metrics, including Peak Signal-to-Noise Ratio (PSNR) \cite{hore2010image}, Learned Perceptual Image Patch Similarity (LPIPS) \cite{zhang2018unreasonable} and Structural Similarity Index Measure (SSIM) \cite{wang2004image}.

\noindent\textbf{Comparison with Baselines.} For quantitative evaluation, we compare our method with existing state-of-the-art (SOTA) approaches:  Gaussian Avatars~\cite{qian2024gaussianavatars} binds 3D Gaussian splats to FLAME mesh triangles with learnable displacements, enabling controllable head synthesis. However, slight misalignments between mesh and images may weaken local details.
Gaussian Head Avatar~\cite{xu2024gaussian} adds an expression-driven MLP deformation field on neutral Gaussians to render ultra-high-fidelity avatars, but relies on dense multi-view capture and complex initialization.
TensorialGaussianAvatar~\cite{wang20253d} uses static triplanes and a lightweight 1D feature line to encode dynamic textures, but relies on multi-view mesh estimation and struggles to align full head textures due to the use of a frozen mesh and limited 1D representation.
MonoGaussianAvatar~\cite{chen2024monogaussianavatar} employs a Gaussian deformation field to reconstruct dynamic avatars from a single-view video, lowering capture requirements; yet the monocular setup lacks geometric constraints, leading to holes and artifacts around teeth and hair during large head motions.
GaussianBlendshapes~\cite{ma20243d} encapsulates high-frequency details into expression-related Gaussian blendshapes, supporting linear mixing and real-time synthesis. However, pure linear interpolation limits generalization to highly complex or non-linear expressions.

\begin{table*}[t]
  \centering
  \small
  \begin{tabular}{c ccc ccc ccc}
    \toprule
    \multirow{2}{*}{Method}
        & \multicolumn{3}{c}{Novel View Synthesis}
    & \multicolumn{3}{c}{Self-Reenactment}
    & \multicolumn{3}{c}{{Self-Reenactment (Novel View)}}\\
    \cmidrule(lr){2-4} \cmidrule(lr){5-7} \cmidrule(lr){8-10} 
    & PSNR$\uparrow$ & SSIM$\uparrow$ & LPIPS$\downarrow$ 
    & PSNR$\uparrow$ & SSIM$\uparrow$ & LPIPS$\downarrow$ 
    & PSNR$\uparrow$ & SSIM$\uparrow$ & LPIPS$\downarrow$  \\
    \midrule

MonoGaussianAvatar     & - & - & -  & - & - & -  & 21.78 & 0.871 & 0.130 \\

GaussianBlendShapes       & - & - & -  & - & - & -   & 24.56 & 0.864 & \textbf{0.088}\\

GaussianAvatars     & 29.71 & 0.939 & 0.079  & 23.58 & 0.891 & 0.093 & 23.01 & 0.900 & 0.100\\

GaussianHeadAvatar  & 26.27 & 0.807 & 0.185  & 19.43 & 0.794 & 0.188 & 19.74 & 0.792 & 0.184\\

TensorialGaussianAvatar     & 26.33 & 0.923 & 0.100 & 23.40  & 0.900 & 0.123 & 23.28 & 0.886 & 0.118 \\

\midrule 

MoGaFace        & \textbf{30.98} & \textbf{0.946} & \textbf{0.068}  & \textbf{24.70} & \textbf{0.900} & \textbf{0.084} & \textbf{24.75} & \textbf{0.902} & 0.090 \\

    \bottomrule

  \end{tabular}
    \caption{Quantitative comparisons with state-of-the-art methods. }

  \label{tab:compared1}
\end{table*}

\noindent\textbf{Implementation Details.}
We evaluate the quality of head avatars with three settings: (1). novel-view synthesis: driving an avatar with head poses and expressions from training sequences and rendering from a held-out viewpoint. (2). self-reenactment: driving an avatar with unseen poses and expressions from a held-out sequence of the same subject and rendering all camera views. (3). cross-identity reenactment: an avatar is driven by the expressions and motions of a different individual.

\begin{figure*}[t]
\centering
\includegraphics[width=0.9\textwidth]{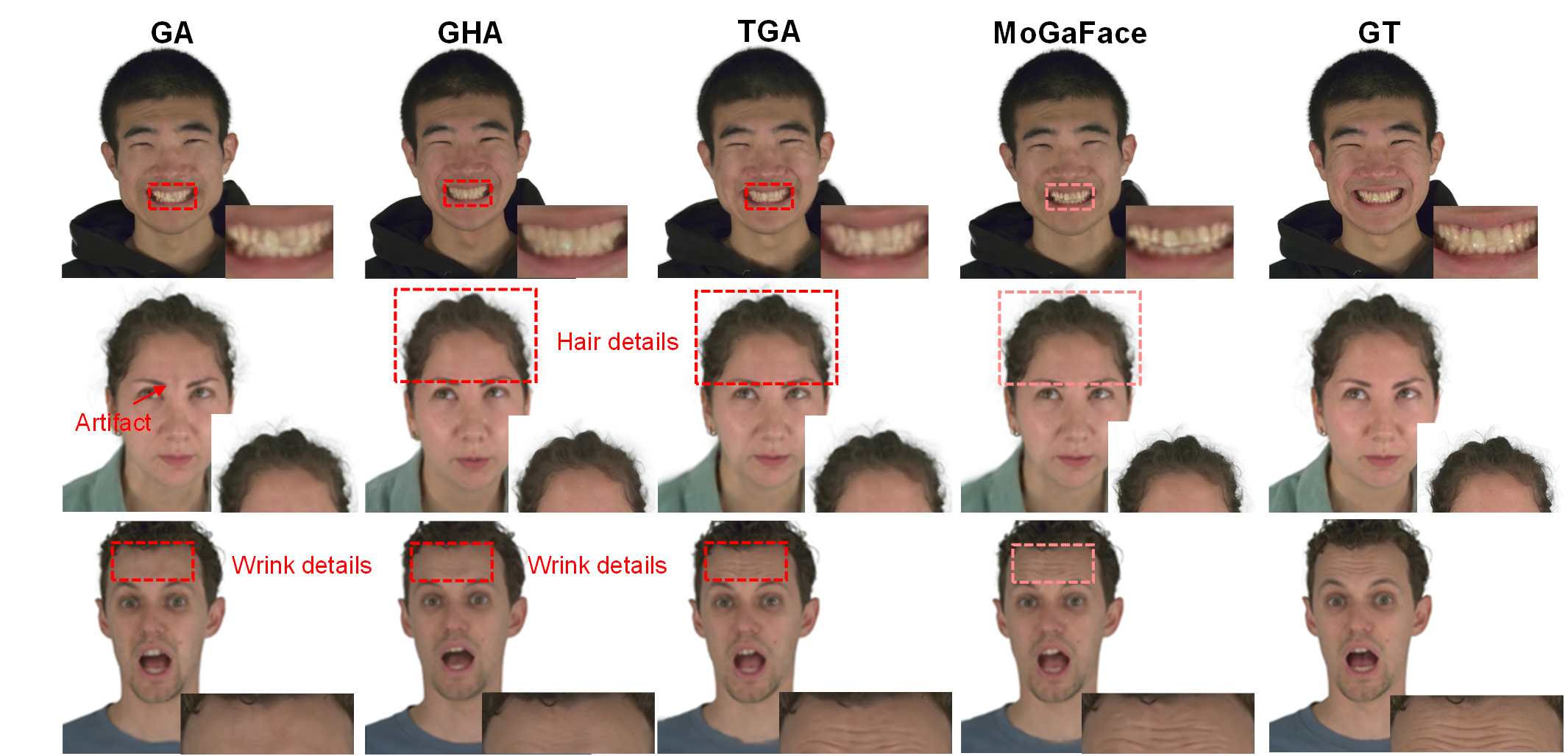} 
\caption{Qualitative visualization with baseline methods on novel view synthesis.}
\label{fig:3}
\end{figure*}

\begin{figure*}[t]
\centering
\includegraphics[width=1\textwidth]{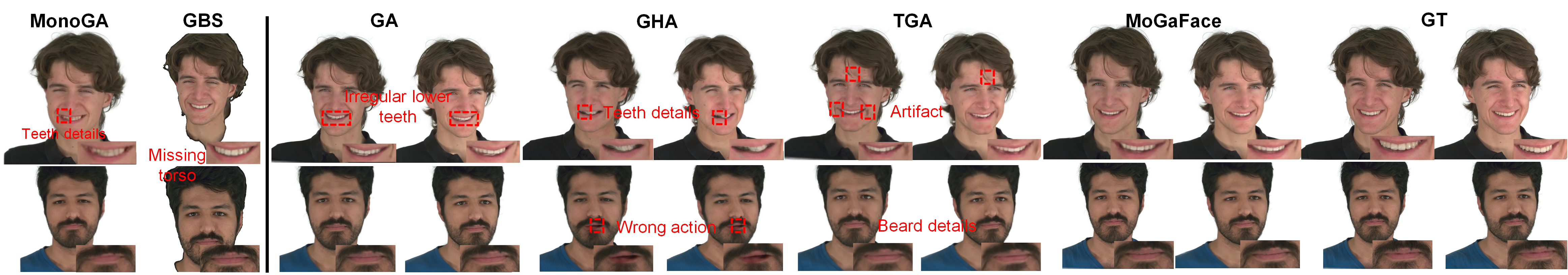} 
\caption{Qualitative visualization with  monocular and multi-view baselines on self-reenactment task.}
\label{fig:self-reenactment}
\end{figure*}

\begin{figure*}[t]
\centering
\includegraphics[width=0.8\textwidth]{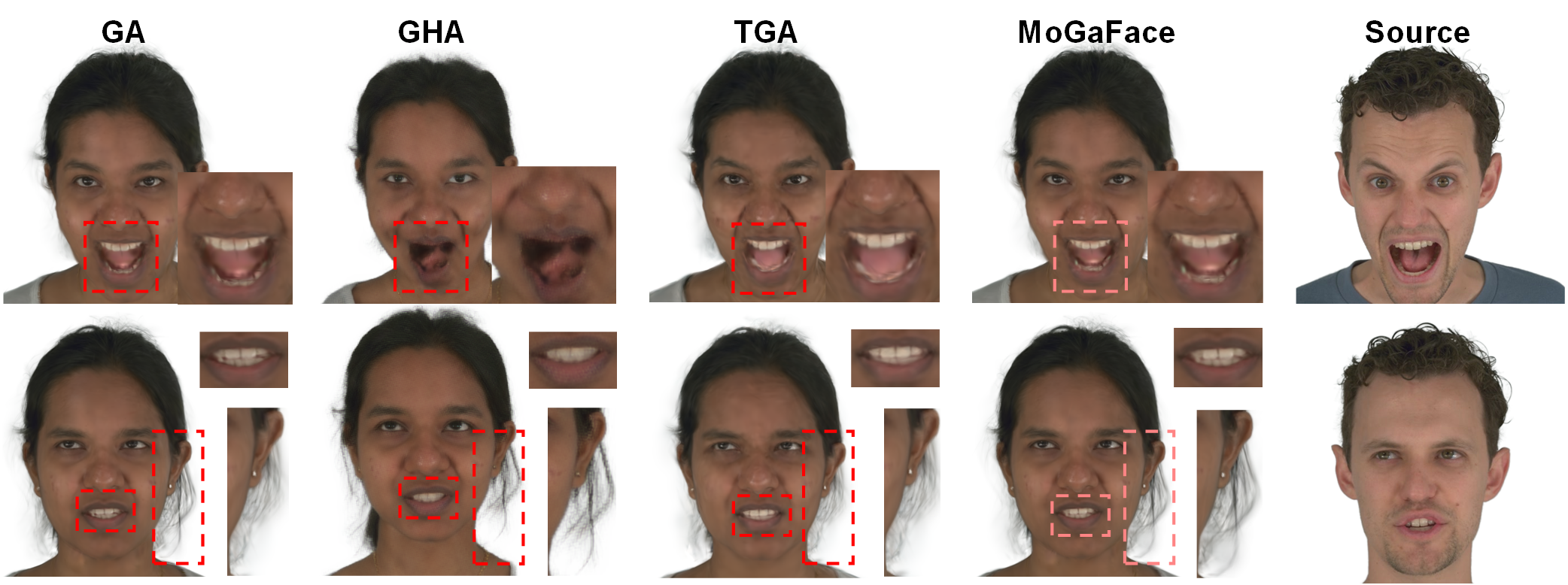} 
\caption{Cross-identity reenactment. Expressions and poses from the source (right) are transferred to the target (left).}
\label{fig:cross}
\end{figure*}

\subsection{Evaluation under Camera-Aware Settings} 
\noindent\textbf{Quantitative Comparisons.} As shown in the Tab.\ref{tab:compared1}, our MoGaFace consistently achieves high PSNR, outperforming SOTA methods by an average of 3.54 in novel view synthesis, and achieving an average improvement of 2.56 in self-reenactment tasks. These results demonstrate the strong 3D reconstruction capability of our method and its superior fidelity at the pixel level. This improvement can be attributed to our dynamic modulation of expression features, which effectively mitigates the fine-grained visual artifacts previously highlighted by GaussianAvatar \cite{qian2024gaussianavatars}. Furthermore, the average gain in SSIM ($+$0.05) and reduction in LPIPS ($-$0.05) further demonstrate MoGaFace’s ability to preserve structural coherence and perceptual quality, validating its effectiveness in maintaining consistency between 3D facial geometry and 2D renderings under multi-view conditions.

\begin{figure}[t]
\centering
\includegraphics[width=1\columnwidth]{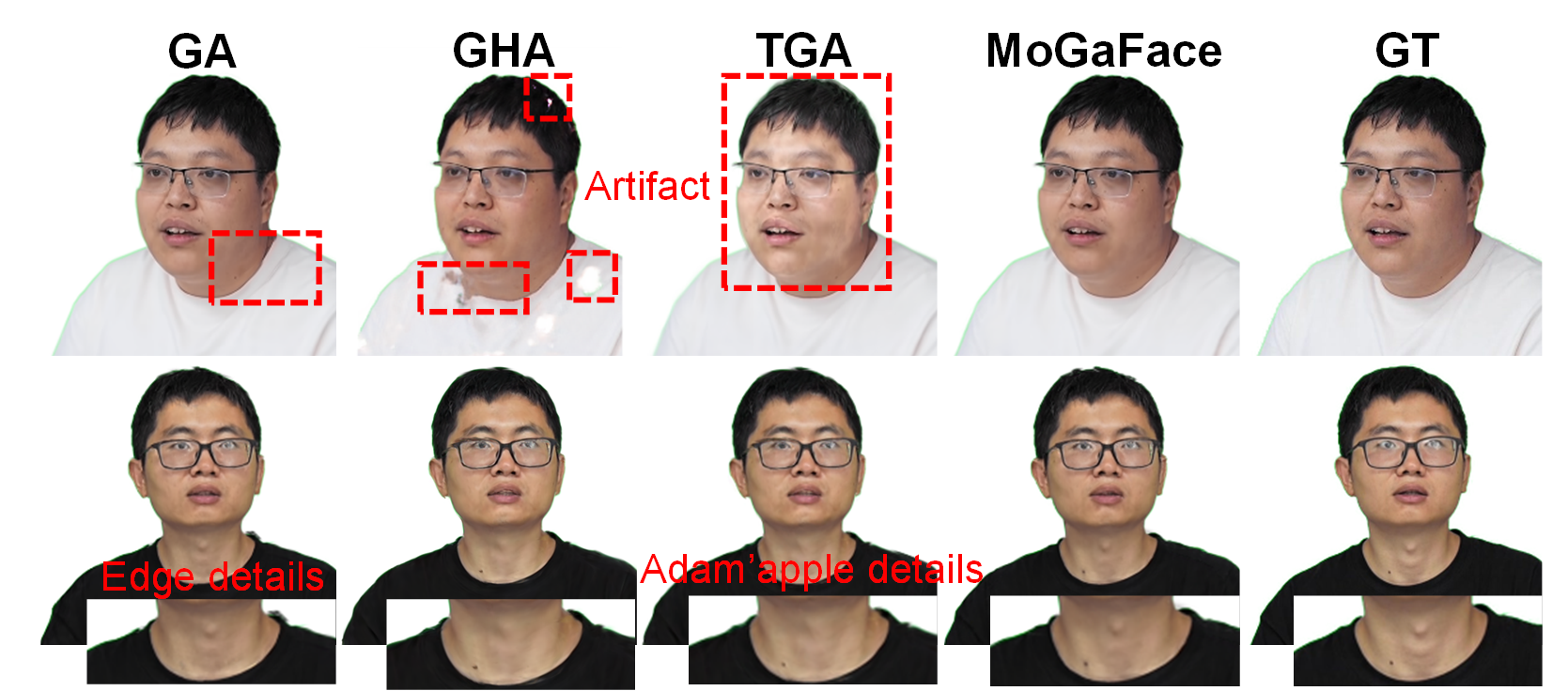} 
\caption{Qualitative visualization with multi-view baselines under camera-free settings.}
\label{fig:cameron_free}
\end{figure}

\noindent\textbf{Qualitative Analysis.} Fig.~\ref{fig:3} and Fig.\ref{fig:self-reenactment} present qualitative comparisons to further evaluate the visual quality and structural accuracy of MoGaFace. In monocular methods (Fig.\ref{fig:self-reenactment}(left), such as MonoGaussianAvatar (MonoGA) and GaussiansBlendShape (GBS), limited visual cues from single-view inputs lead to the loss of fine-grained details, especially around the eyes and mouth, and often cause rendering failures under large facial motions. In multi-view methods (Fig.\ref{fig:3} and Fig.\ref{fig:self-reenactment}(right), the additional spatial context provided by multiple views enables improved reconstruction quality. However, GaussianAvatar (GA) and GaussianHeadAvatar (GHA) suffer from inconsistencies between the 3D geometry and the resulting 2D images, leading to rendering artifacts and blank regions. Moreover, they fail to recover high-fidelity textures, particularly in low-texture areas. TensorialGaussianAvatar (TGA) attempts to embed textures using 1D blendshape features, but the sparsity of these features and their reliance on imprecise mesh geometry result in only shallow and less realistic texture reconstructions. In contrast,  MoGaFace effectively preserves both geometric and appearance consistency across views, benefiting from a multi-view-consistent expression correction strategy and latent-guided texture rendering, which together enhance cross-view fidelity and perceptual realism. Fig.~\ref{fig:cross} illustrates the performance of MoGaFace in the cross-reenactment task, highlighting its ability to generate distinct details while preserving identity.

\begin{table}[htbp]
  \centering
  \small
  \begin{tabular}{c ccc}
    \toprule
    \multirow{2}{*}{Method}
    & \multicolumn{3}{c}{Self-Reenactment} \\
    \cmidrule(lr){2-4} 
    & PSNR$\uparrow$  & SSIM$\uparrow$ & LPIPS$\downarrow$ \\
    \midrule

GaussianAvatars     & 26.83  & 0.938  & 0.034  \\

GaussianHeadAvatar & 21.29  & 0.829 & 0.095   \\

TensorialGaussianAvatar   & 27.11  & 0.941 & 0.055   \\

\midrule 
MoGaFace    & \textbf{28.47}  & \textbf{0.943} & \textbf{0.031}    \\
\bottomrule
  \end{tabular}
    \caption{Quantitative evaluation with multi-view baselines under camera-free settings.}

  \label{tab:ours_data}
\end{table}

\subsection{Evaluation under Camera-Free Settings}
To further evaluate the robustness and practical applicability of MoGaFace, we conduct experiments under camera-free settings using two self-recorded portrait videos without ground-truth camera extrinsics. This setup better reflects real-world scenarios such as mobile or in-the-wild applications, where multi-view calibration is often infeasible and camera poses must be estimated from weakly constrained multi-view inputs to enable scalable deployment. As both Gaussian Splatting and FLAME tracking rely on known camera parameters, we first estimate them using a monocular method\cite{zielonka2022towards}, which inevitably introduces errors. All baselines are evaluated using these predicted parameters for both mesh fitting and Gaussian rendering.

\noindent\textbf{Quantitative Comparisons.} The quantitative results under this setting are presented in Tab. \ref{tab:ours_data}. Inaccurate camera pose estimation often causes geometric distortions in the predicted FLAME mesh, resulting in misalignment between 3D geometry and 2D images. As a result, existing multi-view reconstruction methods experience a noticeable drop in performance, with significantly lower scores across all evaluation metrics. In contrast, MoGaFace incorporates a correction strategy that dynamically refines the FLAME mesh using view-aware cues, ensuring improved 3D–2D consistency even without ground-truth camera parameters. Furthermore, by leveraging latent image features for guided texture refinement, MoGaFace enhances the quality of high-frequency details such as facial contours and skin textures. As a result, it achieves strong performance—PSNR up by 3.39, SSIM by 0.04, and LPIPS down by 0.03—showing robustness to mesh noise and adaptability to camera-free scenarios.

\noindent\textbf{Qualitative Analysis.} The qualitative visualization in Fig. \ref{fig:cameron_free} further illustrate the advantages of MoGaFace. The baseline model with frozen mesh during Gaussian rendering suffers from large-scale facial and body artifacts due to camera parameters errors, and also exhibits clothing boundary leakage, distorted Adam’s apple, and poor shading detail. In contrast, MoGaFace produces sharper and more structurally consistent renderings, demonstrating its effectiveness in recovering high-fidelity geometry and texture under camera-free conditions.
\begin{table}[t]
  \centering
  \small
  \begin{tabular}{c cc cc}
    \toprule
    \multirow{2}{*}{Method}
    & \multicolumn{2}{c}{Novel View Synthesis}
    & \multicolumn{2}{c}{Self-Reenactment} \\
    \cmidrule(lr){2-3} \cmidrule(lr){4-5}
    & PSNR$\uparrow$  & LPIPS$\downarrow$ & PSNR  & LPIPS \\
    \midrule

 Freeze     & 27.49  & 0.121  & 24.35  & 0.134 \\

Single-view &28.71   & 0.080 & 24.42  & 0.105  \\

Multi-view    & \textbf{30.31}  & \textbf{0.079} & \textbf{24.56} & \textbf{0.102}  \\
    \bottomrule
  \end{tabular}
    \caption{Quantitative results of expression correction under different view settings.}

  \label{tab:dynamic_expr}
\end{table}

\subsection{Ablation Study}
We conduct ablation experiments on subject $\#$074 to evaluate the contributions of the three core components of our method: Expression-Aware Dynamic Correction, Momentum-Guided Expression Representation, and Latent-Guided Texture Attention. Each module is evaluated on novel view synthesis and self-reenactment tasks to assess its individual contribution. SSIM results are included in the supplementary due to space limitations.

\noindent\textbf{Expression-Aware Dynamic Correction.} The 2D image features provide essential cues to guide the correction of the FLAME mesh geometry during rendering. However, since FLAME expression coefficients govern the entire head geometry, the limited information from a single image is often insufficient to achieve accurate expression correction. As shown in Table~\ref{tab:dynamic_expr}, using multi-view input for dynamic correction achieves the best performance, with PSNR improved by 2.82 and LPIPS reduced by 0.039 on novel view synthesis, and PSNR improved by 0.21 and LPIPS reduced by 0.032 on self-reenactment compared to the Freeze baseline. In contrast, single-view input results in a noticeable drop, though it still outperforms the Freeze setting, where FLAME expression parameters remain fixed during rendering.

\begin{table}[t]
  \centering
  \small
  \begin{tabular}{c cc cc}
    \toprule
        \multirow{2}{*}{Method}
    & \multicolumn{2}{c}{Novel View Synthesis}
    & \multicolumn{2}{c}{Self-Reenactment} \\
    \cmidrule(lr){2-3} \cmidrule(lr){4-5}
    & PSNR$\uparrow$  & LPIPS$\downarrow$ & PSNR  & LPIPS \\
    \midrule

 \makecell{Multi-view-O}    & 30.31  & \textbf{0.082}  & 24.56  & \textbf{0.102} \\

\makecell{Multi-view-M} & 30.43  & 0.091 & 24.60  & 0.114  \\

\makecell{Multi-view-T}     & \textbf{31.17}  & 0.088 &\textbf{24.63}  & 0.104  \\
    \bottomrule
  \end{tabular}
    \caption{Quantitative results of momentum-guided expression representation and latent-guided texture attention under multi-view settings.}

  \label{tab:bankAndtexture_ab}
\end{table}

\noindent\textbf{Momentum-Guided Expression Representation.} The ablation on Expression-Aware Dynamic Correction verifies the effectiveness of multi-view inputs for expression refinement. However, independently fitted FLAME meshes across views per frame cause inconsistent 3D geometry. To address this, MoGaFace introduces Momentum-Guided Expression Representation for temporal and cross-view consistency.

\begin{figure}[t]
\centering
\includegraphics[width=1\columnwidth]{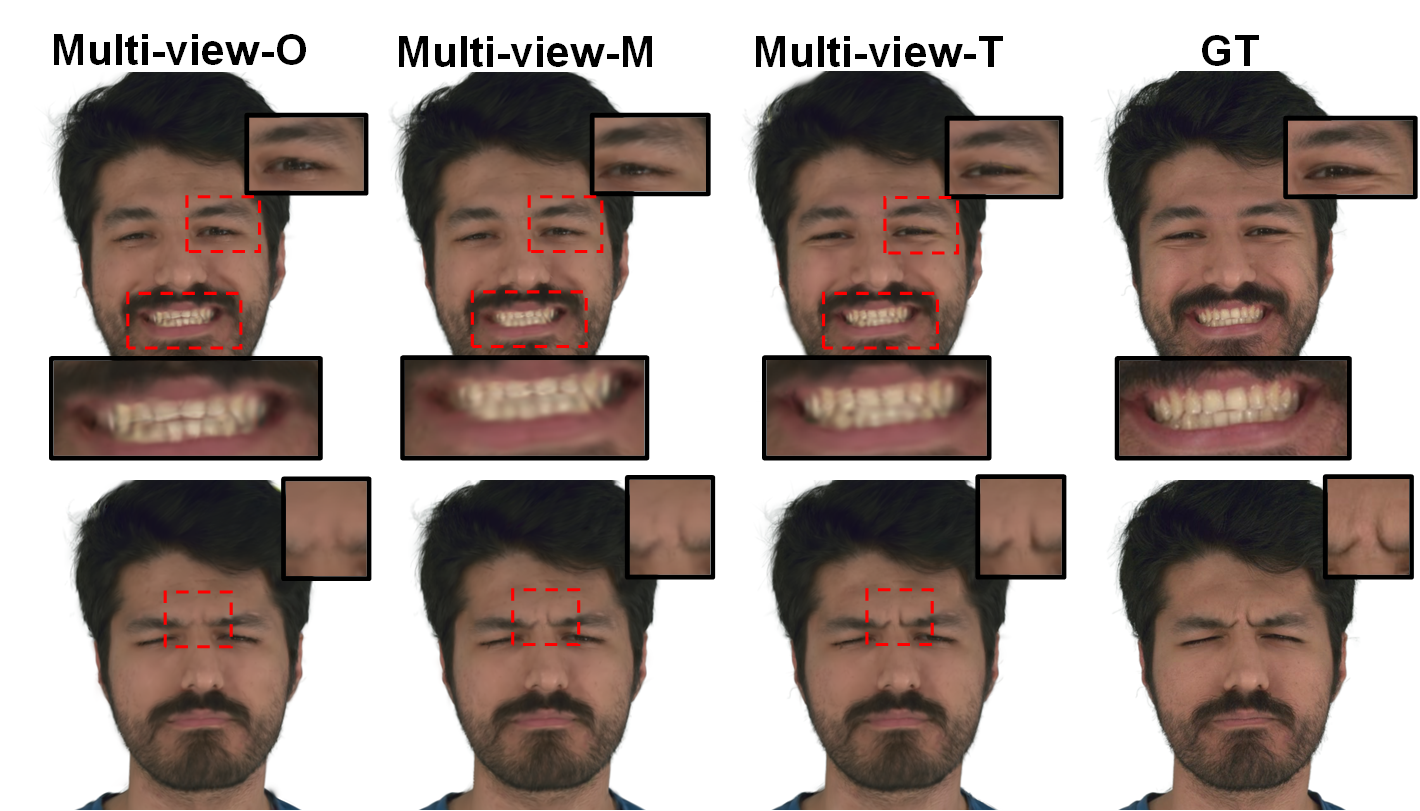} %
\caption{Qualitative visualization of momentum-guided expression representation and latent-guided texture attention under multi-view settings. }
    \label{fig:figure4}
\end{figure}

On top of multi-view dynamic correction (Multi-view-O), we integrate the Momentum-Guided Expression Representation, referred to as Multi-view-M, to enforce a shared FLAME mesh across views at each frame. As shown in Table~\ref{tab:bankAndtexture_ab}, this yields a PSNR gain of 0.12 in novel view synthesis and 0.04 in self-reenactment, while LPIPS slightly increases due to momentum averaging, leading to minor detail loss (e.g., wrinkles, teeth), as shown in Fig. \ref{fig:figure4}.

\noindent\textbf{Latent-Guided Texture Attention.} To improve mesh texture quality, MoGaFace employs The Latent-Guided Texture Attention module to modulate Gaussian properties. To validate its effectiveness, this module is added to the Multi-view-Only setting (Multi-view-O), forming Multi-view-T for fine-grained texture enhancement. As shown in Tab.~\ref{tab:bankAndtexture_ab}, this module improves PSNR by 0.86 in novel view synthesis and 0.07 in self-reenactment. Fig.\ref{fig:figure4} further illustrates qualitative improvements in fine details such as teeth and eye wrinkles.

\section{Conclusion}
In this work, we present MoGaFace, a novel 3D head avatar modeling framework that enables high-fidelity and expressive avatar reconstruction under both camera-aware and camera-free conditions. Unlike prior methods that rely on fixed mesh estimation and suffer from cross-view inconsistencies, MoGaFace integrates multi-view image cues throughout the rendering process to dynamically refine facial expressions and textures. We introduce three key components to address the limitations of prior 3DGS-based avatar systems: (1) Expression-Aware Dynamic Correction, which leverages multi-view information to improve the alignment between 2D image and 3D mesh during rendering; (2) Momentum-Guided Expression Representation, which ensures temporal and cross-view consistency of FLAME expression parameters by maintaining a momentum-updated expression bank; and (3) Latent-Guided Texture Attention, which adaptively incorporates view-aware features to enhance  fine-grained detail synthesis.  Extensive experiments under both standard and uncalibrated settings demonstrate that MoGaFace achieves superior performance in terms of geometric consistency, perceptual quality, and generalization to unseen views and expressions. Moreover, its ability to operate without ground-truth camera parameters significantly enhances its applicability to mobile and in-the-wild scenarios. We believe MoGaFace offers a promising step toward robust and expressive head avatar generation in real-world applications.

\bibliography{aaai22}

\begin{thebibliography}{43}
\providecommand{\natexlab}[1]{#1}

\bibitem[{Aneja et~al.(2024)Aneja, Sevastopolsky, Kirschstein, Thies, Dai, and Nie{\ss}ner}]{aneja2024gaussianspeech}
Aneja, S.; Sevastopolsky, A.; Kirschstein, T.; Thies, J.; Dai, A.; and Nie{\ss}ner, M. 2024.
\newblock Gaussianspeech: Audio-driven gaussian avatars.
\newblock \emph{arXiv preprint arXiv:2411.18675}.

\bibitem[{Arampatzakis et~al.(2023)Arampatzakis, Pavlidis, Mitianoudis, and Papamarkos}]{arampatzakis2023monocular}
Arampatzakis, V.; Pavlidis, G.; Mitianoudis, N.; and Papamarkos, N. 2023.
\newblock Monocular depth estimation: A thorough review.
\newblock \emph{IEEE Transactions on Pattern Analysis and Machine Intelligence}, 46(4): 2396--2414.

\bibitem[{Barron et~al.(2021)Barron, Mildenhall, Tancik, Hedman, Martin-Brualla, and Srinivasan}]{barron2021mip}
Barron, J.~T.; Mildenhall, B.; Tancik, M.; Hedman, P.; Martin-Brualla, R.; and Srinivasan, P.~P. 2021.
\newblock Mip-nerf: A multiscale representation for anti-aliasing neural radiance fields.
\newblock In \emph{Proceedings of the IEEE/CVF international conference on computer vision}, 5855--5864.

\bibitem[{Beeler et~al.(2010)Beeler, Bickel, Beardsley, Sumner, and Gross}]{beeler2010high}
Beeler, T.; Bickel, B.; Beardsley, P.; Sumner, B.; and Gross, M. 2010.
\newblock High-quality single-shot capture of facial geometry.
\newblock In \emph{ACM SIGGRAPH 2010 papers}, 1--9.

\bibitem[{Chen et~al.(2022)Chen, Xu, Geiger, Yu, and Su}]{chen2022tensorf}
Chen, A.; Xu, Z.; Geiger, A.; Yu, J.; and Su, H. 2022.
\newblock Tensorf: Tensorial radiance fields.
\newblock In \emph{European conference on computer vision}, 333--350. Springer.

\bibitem[{Chen et~al.(2024{\natexlab{a}})Chen, Wang, Li, Xiao, Zhang, Yao, and Liu}]{chen2024monogaussianavatar}
Chen, Y.; Wang, L.; Li, Q.; Xiao, H.; Zhang, S.; Yao, H.; and Liu, Y. 2024{\natexlab{a}}.
\newblock Monogaussianavatar: Monocular gaussian point-based head avatar.
\newblock In \emph{ACM SIGGRAPH 2024 Conference Papers}, 1--9.

\bibitem[{Chen et~al.(2024{\natexlab{b}})Chen, Xu, Zheng, Zhuang, Pollefeys, Geiger, Cham, and Cai}]{chen2024mvsplat}
Chen, Y.; Xu, H.; Zheng, C.; Zhuang, B.; Pollefeys, M.; Geiger, A.; Cham, T.-J.; and Cai, J. 2024{\natexlab{b}}.
\newblock Mvsplat: Efficient 3d gaussian splatting from sparse multi-view images.
\newblock In \emph{European Conference on Computer Vision}, 370--386. Springer.

\bibitem[{Fei et~al.(2024)Fei, Xu, Zhang, Zhou, Yang, and He}]{fei20243d}
Fei, B.; Xu, J.; Zhang, R.; Zhou, Q.; Yang, W.; and He, Y. 2024.
\newblock 3d gaussian splatting as new era: A survey.
\newblock \emph{IEEE Transactions on Visualization and Computer Graphics}.

\bibitem[{Feng et~al.(2021)Feng, Feng, Black, and Bolkart}]{feng2021learning}
Feng, Y.; Feng, H.; Black, M.~J.; and Bolkart, T. 2021.
\newblock Learning an animatable detailed 3D face model from in-the-wild images.
\newblock \emph{ACM Transactions on Graphics (ToG)}, 40(4): 1--13.

\bibitem[{Gao et~al.(2022)Gao, Zhong, Xiang, Hong, Guo, and Zhang}]{gao2022reconstructing}
Gao, X.; Zhong, C.; Xiang, J.; Hong, Y.; Guo, Y.; and Zhang, J. 2022.
\newblock Reconstructing personalized semantic facial nerf models from monocular video.
\newblock \emph{ACM Transactions on Graphics (TOG)}, 41(6): 1--12.

\bibitem[{He et~al.(2020)He, Fan, Wu, Xie, and Girshick}]{he2020momentum}
He, K.; Fan, H.; Wu, Y.; Xie, S.; and Girshick, R. 2020.
\newblock Momentum contrast for unsupervised visual representation learning.
\newblock In \emph{Proceedings of the IEEE/CVF conference on computer vision and pattern recognition}, 9729--9738.

\bibitem[{Hore and Ziou(2010)}]{hore2010image}
Hore, A.; and Ziou, D. 2010.
\newblock Image quality metrics: PSNR vs. SSIM.
\newblock In \emph{2010 20th international conference on pattern recognition}, 2366--2369. IEEE.

\bibitem[{Kerbl et~al.(2023)Kerbl, Kopanas, Leimk{\"u}hler, and Drettakis}]{kerbl20233d}
Kerbl, B.; Kopanas, G.; Leimk{\"u}hler, T.; and Drettakis, G. 2023.
\newblock 3D Gaussian splatting for real-time radiance field rendering.
\newblock \emph{ACM Trans. Graph.}, 42(4): 139--1.

\bibitem[{Kirschstein et~al.(2023)Kirschstein, Qian, Giebenhain, Walter, and Nie{\ss}ner}]{kirschstein2023nersemble}
Kirschstein, T.; Qian, S.; Giebenhain, S.; Walter, T.; and Nie{\ss}ner, M. 2023.
\newblock Nersemble: Multi-view radiance field reconstruction of human heads.
\newblock \emph{ACM Transactions on Graphics (TOG)}, 42(4): 1--14.

\bibitem[{Kwon et~al.(2024)Kwon, Fang, Lu, Dong, Zhang, Carrasco, Mosella-Montoro, Xu, Takagi, Kim et~al.}]{kwon2024generalizable}
Kwon, Y.; Fang, B.; Lu, Y.; Dong, H.; Zhang, C.; Carrasco, F.~V.; Mosella-Montoro, A.; Xu, J.; Takagi, S.; Kim, D.; et~al. 2024.
\newblock Generalizable human gaussians for sparse view synthesis.
\newblock In \emph{European Conference on Computer Vision}, 451--468. Springer.

\bibitem[{Li et~al.(2024)Li, Zhang, Bai, Zheng, Ning, Zhou, and Gu}]{li2024talkinggaussian}
Li, J.; Zhang, J.; Bai, X.; Zheng, J.; Ning, X.; Zhou, J.; and Gu, L. 2024.
\newblock Talkinggaussian: Structure-persistent 3d talking head synthesis via gaussian splatting.
\newblock In \emph{European Conference on Computer Vision}, 127--145. Springer.

\bibitem[{Li et~al.(2017)Li, Bolkart, Black, Li, and Romero}]{li2017learning}
Li, T.; Bolkart, T.; Black, M.~J.; Li, H.; and Romero, J. 2017.
\newblock Learning a model of facial shape and expression from 4D scans.
\newblock \emph{ACM Trans. Graph.}, 36(6): 194--1.

\bibitem[{Liu et~al.(2025{\natexlab{a}})Liu, Liu, Hu, Du, Li, Bao, and Wang}]{liu2025review}
Liu, H.; Liu, B.; Hu, Q.; Du, P.; Li, J.; Bao, Y.; and Wang, F. 2025{\natexlab{a}}.
\newblock A review on 3D Gaussian splatting for sparse view reconstruction.
\newblock \emph{Artificial Intelligence Review}, 58(7): 215.

\bibitem[{Liu et~al.(2024{\natexlab{a}})Liu, Wang, Hu, Shen, Ye, Zang, Cao, Li, and Liu}]{liu2024mvsgaussian}
Liu, T.; Wang, G.; Hu, S.; Shen, L.; Ye, X.; Zang, Y.; Cao, Z.; Li, W.; and Liu, Z. 2024{\natexlab{a}}.
\newblock Mvsgaussian: Fast generalizable gaussian splatting reconstruction from multi-view stereo.
\newblock In \emph{European Conference on Computer Vision}, 37--53. Springer.

\bibitem[{Liu et~al.(2024{\natexlab{b}})Liu, Luo, Fan, Wang, Peng, and Zhang}]{liu2024citygaussian}
Liu, Y.; Luo, C.; Fan, L.; Wang, N.; Peng, J.; and Zhang, Z. 2024{\natexlab{b}}.
\newblock Citygaussian: Real-time high-quality large-scale scene rendering with gaussians.
\newblock In \emph{European Conference on Computer Vision}, 265--282. Springer.

\bibitem[{Liu et~al.(2025{\natexlab{b}})Liu, Xu, Guo, Wang, Wang, Tan, and Liu}]{liu2025syncanimation}
Liu, Y.; Xu, S.; Guo, J.; Wang, D.; Wang, Z.; Tan, X.; and Liu, X. 2025{\natexlab{b}}.
\newblock SyncAnimation: A Real-Time End-to-End Framework for Audio-Driven Human Pose and Talking Head Animation.
\newblock \emph{arXiv preprint arXiv:2501.14646}.

\bibitem[{Lombardi et~al.(2021)Lombardi, Simon, Schwartz, Zollhoefer, Sheikh, and Saragih}]{lombardi2021mixture}
Lombardi, S.; Simon, T.; Schwartz, G.; Zollhoefer, M.; Sheikh, Y.; and Saragih, J. 2021.
\newblock Mixture of volumetric primitives for efficient neural rendering.
\newblock \emph{ACM Transactions on Graphics (ToG)}, 40(4): 1--13.

\bibitem[{Ma et~al.(2024)Ma, Weng, Shao, and Zhou}]{ma20243d}
Ma, S.; Weng, Y.; Shao, T.; and Zhou, K. 2024.
\newblock 3d gaussian blendshapes for head avatar animation.
\newblock In \emph{ACM SIGGRAPH 2024 Conference Papers}, 1--10.

\bibitem[{Mildenhall et~al.(2021)Mildenhall, Srinivasan, Tancik, Barron, Ramamoorthi, and Ng}]{mildenhall2021nerf}
Mildenhall, B.; Srinivasan, P.~P.; Tancik, M.; Barron, J.~T.; Ramamoorthi, R.; and Ng, R. 2021.
\newblock Nerf: Representing scenes as neural radiance fields for view synthesis.
\newblock \emph{Communications of the ACM}, 65(1): 99--106.

\bibitem[{M{\"u}ller et~al.(2022)M{\"u}ller, Evans, Schied, and Keller}]{muller2022instant}
M{\"u}ller, T.; Evans, A.; Schied, C.; and Keller, A. 2022.
\newblock Instant neural graphics primitives with a multiresolution hash encoding.
\newblock \emph{ACM transactions on graphics (TOG)}, 41(4): 1--15.

\bibitem[{Peng et~al.(2024)Peng, Hu, Shi, Zhu, Zhang, Zhao, He, Liu, and Fan}]{peng2024synctalk}
Peng, Z.; Hu, W.; Shi, Y.; Zhu, X.; Zhang, X.; Zhao, H.; He, J.; Liu, H.; and Fan, Z. 2024.
\newblock Synctalk: The devil is in the synchronization for talking head synthesis.
\newblock In \emph{Proceedings of the IEEE/CVF Conference on Computer Vision and Pattern Recognition}, 666--676.

\bibitem[{Qian et~al.(2024)Qian, Kirschstein, Schoneveld, Davoli, Giebenhain, and Nie{\ss}ner}]{qian2024gaussianavatars}
Qian, S.; Kirschstein, T.; Schoneveld, L.; Davoli, D.; Giebenhain, S.; and Nie{\ss}ner, M. 2024.
\newblock Gaussianavatars: Photorealistic head avatars with rigged 3d gaussians.
\newblock In \emph{Proceedings of the IEEE/CVF Conference on Computer Vision and Pattern Recognition}, 20299--20309.

\bibitem[{Qiu et~al.(2025)Qiu, Zhu, Zuo, Gu, Dong, Zhang, Xu, Li, Yuan, Bo et~al.}]{qiu2025anigs}
Qiu, L.; Zhu, S.; Zuo, Q.; Gu, X.; Dong, Y.; Zhang, J.; Xu, C.; Li, Z.; Yuan, W.; Bo, L.; et~al. 2025.
\newblock Anigs: Animatable gaussian avatar from a single image with inconsistent gaussian reconstruction.
\newblock In \emph{Proceedings of the Computer Vision and Pattern Recognition Conference}, 21148--21158.

\bibitem[{Tang and Cham(2024)}]{tang20243igs}
Tang, Z.~J.; and Cham, T.-J. 2024.
\newblock 3igs: Factorised tensorial illumination for 3d gaussian splatting.
\newblock In \emph{European Conference on Computer Vision}, 143--159. Springer.

\bibitem[{Thies et~al.(2016)Thies, Zollhofer, Stamminger, Theobalt, and Nie{\ss}ner}]{thies2016face2face}
Thies, J.; Zollhofer, M.; Stamminger, M.; Theobalt, C.; and Nie{\ss}ner, M. 2016.
\newblock Face2face: Real-time face capture and reenactment of rgb videos.
\newblock In \emph{Proceedings of the IEEE conference on computer vision and pattern recognition}, 2387--2395.

\bibitem[{Wang et~al.(2025{\natexlab{a}})Wang, Xie, Li, Xu, Pun, and Gao}]{wang2025gaussianhead}
Wang, J.; Xie, J.-C.; Li, X.; Xu, F.; Pun, C.-M.; and Gao, H. 2025{\natexlab{a}}.
\newblock Gaussianhead: High-fidelity head avatars with learnable gaussian derivation.
\newblock \emph{IEEE Transactions on Visualization and Computer Graphics}.

\bibitem[{Wang et~al.(2025{\natexlab{b}})Wang, Wang, Yi, Fan, Hu, Zhu, and Ma}]{wang20253d}
Wang, Y.; Wang, X.; Yi, R.; Fan, Y.; Hu, J.; Zhu, J.; and Ma, L. 2025{\natexlab{b}}.
\newblock 3D Gaussian Head Avatars with Expressive Dynamic Appearances by Compact Tensorial Representations.
\newblock In \emph{Proceedings of the Computer Vision and Pattern Recognition Conference}, 21117--21126.

\bibitem[{Wang et~al.(2004)Wang, Bovik, Sheikh, and Simoncelli}]{wang2004image}
Wang, Z.; Bovik, A.~C.; Sheikh, H.~R.; and Simoncelli, E.~P. 2004.
\newblock Image quality assessment: from error visibility to structural similarity.
\newblock \emph{IEEE transactions on image processing}, 13(4): 600--612.

\bibitem[{Wei et~al.(2025)Wei, Zhang, Zhang, Shao, and Lu}]{wei2025pcr}
Wei, Y.; Zhang, J.; Zhang, X.; Shao, L.; and Lu, S. 2025.
\newblock PCR-GS: COLMAP-Free 3D Gaussian Splatting via Pose Co-Regularizations.
\newblock \emph{arXiv preprint arXiv:2507.13891}.

\bibitem[{Wu et~al.(2018)Wu, Xiong, Yu, and Lin}]{wu2018unsupervised}
Wu, Z.; Xiong, Y.; Yu, S.~X.; and Lin, D. 2018.
\newblock Unsupervised feature learning via non-parametric instance discrimination.
\newblock In \emph{Proceedings of the IEEE conference on computer vision and pattern recognition}, 3733--3742.

\bibitem[{Xie et~al.(2025)Xie, Chen, Xu, Xie, Jin, Shen, Peng, Bao, and Zhou}]{xie2025envgs}
Xie, T.; Chen, X.; Xu, Z.; Xie, Y.; Jin, Y.; Shen, Y.; Peng, S.; Bao, H.; and Zhou, X. 2025.
\newblock Envgs: Modeling view-dependent appearance with environment gaussian.
\newblock In \emph{Proceedings of the Computer Vision and Pattern Recognition Conference}, 5742--5751.

\bibitem[{Xu et~al.(2024)Xu, Chen, Li, Zhang, Wang, Zheng, and Liu}]{xu2024gaussian}
Xu, Y.; Chen, B.; Li, Z.; Zhang, H.; Wang, L.; Zheng, Z.; and Liu, Y. 2024.
\newblock Gaussian head avatar: Ultra high-fidelity head avatar via dynamic gaussians.
\newblock In \emph{Proceedings of the IEEE/CVF conference on computer vision and pattern recognition}, 1931--1941.

\bibitem[{Xu et~al.(2023)Xu, Zhang, Wang, Zhao, Huang, Qi, and Liu}]{xu2023latentavatar}
Xu, Y.; Zhang, H.; Wang, L.; Zhao, X.; Huang, H.; Qi, G.; and Liu, Y. 2023.
\newblock Latentavatar: Learning latent expression code for expressive neural head avatar.
\newblock In \emph{ACM SIGGRAPH 2023 Conference Proceedings}, 1--10.

\bibitem[{Yu et~al.(2024)Yu, Qu, Yu, Chen, Jiang, Chen, Zhang, Xu, Wu, Lv et~al.}]{yu2024gaussiantalker}
Yu, H.; Qu, Z.; Yu, Q.; Chen, J.; Jiang, Z.; Chen, Z.; Zhang, S.; Xu, J.; Wu, F.; Lv, C.; et~al. 2024.
\newblock Gaussiantalker: Speaker-specific talking head synthesis via 3d gaussian splatting.
\newblock In \emph{Proceedings of the 32nd ACM International Conference on Multimedia}, 3548--3557.

\bibitem[{Zhang, Chen, and Wang(2023)}]{zhang2023explicifying}
Zhang, R.; Chen, J.; and Wang, Q. 2023.
\newblock Explicifying neural implicit fields for efficient dynamic human avatar modeling via a neural explicit surface.
\newblock In \emph{Proceedings of the 31st ACM International Conference on Multimedia}, 1955--1963.

\bibitem[{Zhang et~al.(2018)Zhang, Isola, Efros, Shechtman, and Wang}]{zhang2018unreasonable}
Zhang, R.; Isola, P.; Efros, A.~A.; Shechtman, E.; and Wang, O. 2018.
\newblock The unreasonable effectiveness of deep features as a perceptual metric.
\newblock In \emph{Proceedings of the IEEE conference on computer vision and pattern recognition}, 586--595.

\bibitem[{Zheng et~al.(2024)Zheng, Zhou, Shao, Liu, Zhang, Nie, and Liu}]{zheng2024gps}
Zheng, S.; Zhou, B.; Shao, R.; Liu, B.; Zhang, S.; Nie, L.; and Liu, Y. 2024.
\newblock Gps-gaussian: Generalizable pixel-wise 3d gaussian splatting for real-time human novel view synthesis.
\newblock In \emph{Proceedings of the IEEE/CVF conference on computer vision and pattern recognition}, 19680--19690.

\bibitem[{Zielonka, Bolkart, and Thies(2022)}]{zielonka2022towards}
Zielonka, W.; Bolkart, T.; and Thies, J. 2022.
\newblock Towards metrical reconstruction of human faces.
\newblock In \emph{European conference on computer vision}, 250--269. Springer.

\end{thebibliography}

\end{document}